\documentclass[letterpaper, 10 pt, conference]{ieeeconf}  

\IEEEoverridecommandlockouts                              

\usepackage{xcolor}
\usepackage{flushend}
\definecolor{green}{RGB}{11,155,13}
\newcommand{\xuesu}[1]{{\textcolor{red}{[xuesu: #1]}}}

\title{\LARGE \bf High-Speed Accurate Robot Control using Learned Forward Kinodynamics and Non-linear Least Squares Optimization}

\author{Pranav Atreya$^{1}$, Haresh Karnan$^{2}$, Kavan Singh Sikand$^{1}$,\\ Xuesu Xiao$^{1}$, Sadegh Rabiee$^{1}$, and Joydeep Biswas$^{1}$
\thanks{$^{1}$The University of Texas at Austin, Department of Computer Science, {\tt\small \{pranavatreya, kvsikand\}@utexas.edu, \{xiao, srabiee, joydeepb\}@cs.utexas.edu}}%
\thanks{$^{2}$The University of Texas at Austin, Department of Mechanical Engineering {\tt\small haresh.miriyala@utexas.edu}}%
}

\usepackage{amsmath}
\usepackage{cite}
\usepackage{caption}
\usepackage{graphicx}
\usepackage{amssymb}
\usepackage{xcolor}
\usepackage{caption}
\usepackage{subcaption}
\usepackage{balance} 
\usepackage{graphicx}
\graphicspath{ {./images/} }
\usepackage{caption}
\usepackage{subcaption}
\usepackage{pgf}
\usepackage {multirow}
\usepackage [english]{babel}
\usepackage{fixltx2e}
\usepackage{setspace}
\usepackage{booktabs}
\usepackage{array}
\usepackage{tabularx}
\usepackage{calc}
\usepackage{makecell}
\usepackage{algorithm}
\usepackage{algpseudocode}
\usepackage{paralist}
\usepackage{ragged2e}
\usepackage{verbatim}

\DeclareMathOperator*{\argmax}{arg\,max}
\DeclareMathOperator*{\argmin}{arg\,min}

\begin{document}

\maketitle
\thispagestyle{empty}
\pagestyle{empty}

\begin{abstract}
Accurate control of robots at high speeds requires a control system that can take into account the kinodynamic interactions of the robot with the environment. Prior works on learning inverse kinodynamic (IKD) models of robots have shown success in capturing the complex kinodynamic effects. However, the types of control problems these approaches can be applied to are limited only to that of following pre-computed kinodynamically feasible trajectories. In this paper we present Optim-FKD, a new formulation for accurate, high-speed robot control that makes use of a learned forward kinodynamic (FKD) model and non-linear least squares optimization. Optim-FKD can be used for accurate, high speed control on \emph{any} control task specifiable by a non-linear least squares objective. Optim-FKD can solve for control objectives such as path following and time-optimal control in real time, without needing access to pre-computed kinodynamically feasible trajectories. We empirically demonstrate these abilities of our approach through experiments on a scale one-tenth autonomous car. Our results show that Optim-FKD can follow desired trajectories more accurately and can find better solutions to optimal control problems than baseline approaches.

\end{abstract}

\section{Introduction and Related Work}

At moderate speeds, pure kinematic models or simplified kinodynamic models are sufficient for point to point motion control, for example, in model predictive control~\cite{tahirovic2010mobile, park2012robot, lorenzetti2019reduced} schemes. Such simplified models assume that robots only operate in a limited subspace of their entire state space, such as low acceleration and speed, minimum wheel slip, negligible tire deformation, and  perfect non-holonomic constraints. 
However, real-world robotic missions such as high-speed off-road driving may entail violations of such simplified kinodynamic models.

Since accurate kinodynamic models are hard to construct analytically, there has recently been considerable interest in applying learning to motion control~\cite{xiao2020motion}. End-to-end learning is the most straightforward way to encapsulate both the model and controller in one function approximator and to train it with data, e.g., learning a neural network using imitation learning~\cite{bojarski2016end, tai2016deep, pfeiffer2017perception, pan2020imitation, xiao2021toward, xiao2021agile, wang2021agile, voila} or reinforcement learning~\cite{tai2017virtual, pfeiffer2018reinforced, chiang2019learning, chen2017decentralized}. 
Such end-to-end approaches suffer from drawbacks such as requiring a significant volume of training data, poor generalizability to unseen scenarios, and most importantly, a lack of safety assurance for navigation, which is commonly provided by classical planning and control algorithms. 

To address such disadvantages of end-to-end learning, hybrid approaches have been introduced that leverage existing robot planning and control methods. For example, Wigness et al.~\cite{wigness2018robot} and Sikand et al.~\cite{sikand2021visual} used imitation learning to learn a cost function for existing navigation controllers to enable adaptive behaviors for semantics. Similar techniques have been applied to learn socially compliant navigation \cite{kim2016socially, kretzschmar2016socially, scand}. Xiao et al.~\cite{xiao2021appl} introduced Adaptive Planner Parameter Learning, which learns to dynamically adjust existing motion planners' parameters to efficiently navigate through different obstacle configurations using teleoperated demonstration~\cite{xiao2020appld}, corrective interventions~\cite{wang2021appli}, evaluative feedback~\cite{wang2021apple}, and reinforcement learning~\cite{xu2021applr}. These approaches focus on using machine learning to enable custom behaviors, such as semantic awareness, social compliance, or smooth obstacle-avoidance. 

Machine learning has also been used for high-speed robot control~\cite{betz2022autonomous, xiao2021toward, viikd}. Model Predictive Path Integral control~\cite{williams2017model} utilizes a classical sampling based controller in an online learning fashion: it learns a sample distribution during online deployment that is likely to generate good samples~\cite{williams2017information}. However, it uses a simple unicycle forward model to predict future paths based on the samples and compensates for such an over-simplified model by a massive number of samples evaluated in parallel on GPUs~\cite{williams2016aggressive}. Brunnbauer et al.~\cite{brunnbauer2021model} reported that model-based deep reinforcement learning substantially outperforms model-free agents with respect to performance, sample efficiency, successful task completion, and generalization in autonomous racing. 
Learning inverse kinodynamics conditioned on inertial observations~\cite{xiao2021learning} has also been shown to be effective for accurate, high-speed, off-road navigation on unstructured terrain.

Another class of approaches apply machine learning to model predictive control~\cite{piche2000nonlinear, kabzan2019learning, hewing2020learning}. These approaches however either use simple learned primitives for modelling dynamics, or are geared primarily towards process control, and have not been demonstrated to be capable of controlling complex robotic systems at high speed in real time. 

Leveraging a hybrid paradigm to address high-speed robot control problems, our approach, which we call Optim-FKD, utilizes non-linear least squares optimization to find optimal control sequences based on the learned model in order to enable high-speed, accurate robot motions. 
The contributions of this paper are summarized as follows:
\begin{enumerate}
        \item A novel formulation for robotic control using a learned forward kinodynamic 
                function and non-linear least squares optimization. We demonstrate that this formulation is easily 
                extensible to a range of control tasks without requiring the retraining of a new 
                forward kinodynamic model. 
        \item A novel learning formulation that enables a highly accurate forward kinodynamic 
                model to be learned. 
        \item A detailed description of the system architecture required to enable the presented approach 
                to run on real robot hardware in real time. 
        \item Empirical results demonstrating that the presented approach outperforms baselines for 
                various robot control tasks. 
\end{enumerate}

\section{Optim-FKD Mathematical Formulation}

We first present a general mathematical formulation for optimal control stated as a non-linear least squares problem using a learned forward kinodynamic function. Next, we demonstrate how such a formulation can be applied to different objectives, including trajectory following and time-optimal control.

\subsection{Preliminaries}


Let $X$ represent the state space of the robot. $X$ consists of configuration space 
variables (such as position and orientation) and dynamics variables (such as linear and angular 
velocity). Let $U$ represent the control space of the robot. Consider a period of time of operation of 
the robot $\Delta t$. It is assumed that controls are executed on the robot in a piecewise constant 
manner. Let $\tau$ be the duration for which a particular constant control is executed. In the time period 
of operation $\Delta t$, the robot will execute $n = \frac{\Delta t}{\tau}$ constant controls. 

To model the response of the robot from the executed controls, we introduce a state transition 
likelihood function $\rho : X \times U^n \times X^n \rightarrow [0, 1]$. $\rho$ takes as input 
the initial state of the robot $x_0 \in X$, a length-$n$ piecewise constant control sequence 
$u_{1:n}$, and a length-$n$ state sequence $x_{1:n}$. Each $x_i \in x_{1:n}$ 
represents the state of the robot at time $i \cdot \tau$. The output of $\rho$ is the probability that 
the state sequence $x_{1:n}$ is observed after executing $u_{1:n}$ beginning from $x_0$. 

We assume that the motion of the robot obeys the Markov property, that is, the probability of reaching 
a state $x_i$ depends only on the previous state $x_{i-1}$ and the constant control executed 
beginning at that previous state $u_i$. This induces a local state transition likelihood function 
$\rho_i(x_{i-1}, u_i, x_i)$ for every $i \in 1 ... n$. 
The probability that $x_{1:n}$ is observed after executing 
$u_{1:n}$ from $x_0$ can be written in terms of $\rho_i$ as \begin{equation}
        \label{markov_eq}
        \rho(x_0, u_{1:n}, x_{1:n}) = \prod_{i=1}^{n}{\rho_i(x_{i-1}, u_i, x_i)} \quad .
\end{equation}
Moreover, the maximum likelihood state sequence $\widehat{x_{1:n}}$ after executing $u_{1:n}$ from $x_0$ is
\begin{align}
        \widehat{x_{1:n}} &= \argmax_{x_{1:n}}{\rho(x_0, u_{1:n}, x_{1:n})} \\
        &= \argmax_{x_{1:n}}{\prod_{i=1}^{n}{\rho_i(x_{i-1}, u_i, x_i)}} \label{max_likelihood_state_seq} \quad.
\end{align}
Breaking down each generative probability $\rho_i$ into discriminative probabilities yields
\begin{align}
    \rho_i(x_{i-1}, u_i, x_i) &= p(x_i | u_i, x_{i-1})p(x_{i-1}|u_i)p(u_i) \label{independence0} \\
    &= p(x_i | u_i, x_{i-1})p(x_{i-1})p(u_i) \quad , \label{independence}
\end{align}
where Eq.~\ref{independence} is obtained by applying the fact that the previous state $x_{i-1}$ is independent from the next control. Plugging Eq.~\ref{independence} in Eq.~\ref{max_likelihood_state_seq} yields $\widehat{x_{1:n}} = \argmax_{x_{1:n}}{\prod_{i=1}^{n}{p(x_i | u_i, x_{i-1})}}$.
We assume $p(x_i | u_i, x_{i-1})$ follows a normal distribution: $p(x_i | u_i, x_{i-1}) \sim \mathcal{N}(\bar{x_i}, \sigma_{x_i})$, and represent the maximum likelihood estimate of $p(x_i | u_i, x_{i-1})$ as the forward kinodynamic 
function $\pi(u_i, x_{i-1}) = \bar{x_i}$. With this definition of $\pi$, equation \ref{max_likelihood_state_seq}
can be rewritten as 
\begin{equation}
        \label{incorporating_fkd_model}
        \widehat{x_{1:n}} = (\pi(u_1, x_0), ..., \pi(u_n, \widehat{x_{n-1}}))
\end{equation}

Next, we show that various 
robot control problems can be expressed as nonlinear least squares optimizations that use the 
forward kinodynamic function $\pi$. 

\subsection{Objective 1: Path Following}
\label{formulation_1}

The problem we consider here is that of following a predefined path as closely as possible. This problem 
becomes noteworthy at high speeds where accurate control of the robot becomes increasingly more 
difficult. 

We are given $x_{1:n}^*$ which describes a path to follow. Following this path as closely 
as possible amounts to solving
\begin{equation}
        u_{1:n}^* = \argmax_{u_{1:n}}{\rho(x_0, u_{1:n}, x_{1:n}^*)} \quad .
\end{equation}
From Eq. \ref{incorporating_fkd_model}, this is equivalent to solving
\begin{equation}
        \label{optim_objective_1}
        u_{1:n}^* = \argmin_{u_{1:n}}{||x_{1:n}^* - \widehat{x_{1:n}}||_2^2} \quad,
\end{equation}
which is a nonlinear least squares problem where each $\widehat{x_i} \in \widehat{x_{1:n}}$
is determined from the forward kinodynamic function $\pi$. 

\subsection{Objective 2: Optimal Connectivity}
\label{formulation_two}
Another variant of the robot control problem that we consider is reaching a goal state $x_f$ from a start state $x_i$ in the shortest amount of time possible.
Problems of this type appear frequently in 
optimal sampling-based motion planning, where algorithms like RRT*~\cite{karaman2011sampling} and BIT*~\cite{gammell2015batch} require a 
steering function that can time-optimally connect arbitrary states. Learned kinodynamic models for such problems have so far only been  studied in simulation~\cite{atreya2022state}.

Consider the maximum likelihood state sequence $\widehat{x_{1:n}}$ from earlier. If we 
wanted the final state of the robot to be as close as possible to the goal state $x_f$, we would 
optimize
\begin{equation}
        u_{1:n}^* = \argmin_{u_{1:n}}{||x_f - \widehat{x_n}||_2^2}
\end{equation}
This formulation however keeps the time that the goal state $x_f$ is reached fixed. Specifically, 
the state $\widehat{x_n}$ is reached after time $n \cdot \tau$. To also minimize the time taken to reach 
the goal, $n$ is introduced as an optimization parameter. This results in the objective function
\begin{equation}
        \label{optim_objective_2}
        u_{1:n}^*, n^* = \argmin_{u_{1:n}, n}{||x_f - \widehat{x_n}||_2^2 + (\alpha(n \cdot \tau))^2}
\end{equation}
where $\alpha$ is a scaling parameter that trades off time to reach the goal and the distance to 
the goal. Like the path following formulation, this formulation is a nonlinear least squares optimization where 
$\widehat{x_n}$ is determined from the forward kinodynamic function $\pi$. 


\section{Forward Kinodynamic Model Learning}
In this section we present how the forward kinodynamic model $\pi$ is learned. Since $\pi$ is an 
integral component to the nonlinear least squares optimizations introduced earlier, it is key that 
$\pi$ models the true forward kinodynamics effectively. We learn $\pi$ for a scale one-tenth 
autonomous robot car. 

\subsection{Dataset Generation}
The FKD model $\pi$ is trained in a supervised manner. We obtain the labeled training data by teleoperating the robot at various speeds and recording at every timestep the state estimates of the robot and the joysticked control commands. This results in a dataset $D$ of trajectories 
$T_1, ..., T_m$ where each trajectory $T_i \in D$ is a tuple of the form $(v_x(t), v_y(t), \omega(t), x(t), 
y(t), \theta(t), \delta(t), \psi(t))$. Here, $v_x(t)$ is the velocity in the $x$-direction, $v_y(t)$ is the 
velocity in the $y$-direction, $\omega(t)$ is the angular velocity, $x(t)$ is the $x$-position, 
$y(t)$ is the $y$-position, $\theta(t)$ is the orientation, $\delta(t)$ is the commanded forward velocity, 
and $\psi(t)$ is the commanded angular velocity. All of these functions are time-dependent 
and are defined in the domain $[0, t_f^{(i)}]$ where $t_f^{(i)}$ is the termination time for trajectory $T_i$.

\subsection{Learning Formulation}

The formulation presented earlier for $\pi$ maps an initial state of the robot and a constant 
control to the most likely immediate next state of the robot. 
Training of an FKD model with such short prediction horizon is challenging in practice. If the model is 
tasked with predicting the state after $\tau$ units of time, where $\tau$ is a small positive number, it can get away with simply predicting 
the current state without incurring much loss. In order to ensure the model's predictions are of high 
quality, the model needs to learn to model the state of the robot after a time period much greater than 
$\tau$. 

Simply increasing $\tau$ 
would not suffice for achieving a longer prediction horizon since that would forego the fine-grained prediction capabilities of the model. We address this problem by training $\pi$ in a recurrent fashion. The basic 
structure of the recurrence formulation is as follows. The model predicts the next state 
$x_i = \pi(u_i, x_{i-1})$. For timestep $i+1$, instead of being given access to the ground truth value of 
$x_i$, the model uses its previous prediction as the starting state: $x_{i+1} = \pi(u_{i+1}, x_i)$. This 
process continues for the number of timesteps in the prediction horizon. 

This simple recurrent approach has a few limitations however. The base timestep duration $\tau$ is 
selected to be small so as to capture minute changes in the state of the robot. In our experiments we 
set $\tau$ to $0.05$ seconds. For the model to predict the state of the robot after time $t_\mathrm{pred}$, 
$\frac{t_\mathrm{pred}}{\tau}$ forward passes through the model are needed. In our experiments, we set 
$t_\mathrm{pred}$ to $3.0$ seconds, requiring $60$ forward passes through the model. Since $\pi$ is to 
be used in a real-time optimization framework, the number of forward passes through $\pi$ need to be limited 
to maintain computational efficiency. We achieve this by introducing a model prediction time $t_\mathrm{model}$. 
The FKD model $\pi$, in one forward pass, outputs the next $\frac{t_\mathrm{model}}{\tau}$ states given the 
next $\frac{t_\mathrm{model}}{\tau}$ controls. It also takes in as input the previous $\frac{t_\mathrm{model}}{\tau}$
states, enabling recurrence. In our experiments $t_\mathrm{model}$ was set to $0.5$ seconds. This approach 
enables both computational efficiency and fine-grained prediction. 

Having motivated the model formulation, we now present the learning objective. For each trajectory $T_i$ 
in our dataset $D$, we evenly sample $k$ starting times $(t_1, ..., t_k)$ from the range 
$[t_\mathrm{model}, t_f^{(i)}-t_\mathrm{pred}]$. For each starting time $t_a \in (t_1, ..., t_k)$, 
we use the model to predict the states at times $t_a + b\tau$ for $b \in [0, 1, ..., 
\frac{t_\mathrm{pred}}{\tau}]$. For brevity, let $S_{t_a}$ be the state variables $v_x(t), v_y(t), 
\omega(t), x(t), y(t)$, and $\theta(t)$ sampled evenly with spacing $\tau$ from the time period 
$[t_a, t_a + t_\mathrm{model}]$. Additionally let $M_{t_a}$ be the control variables $\delta(t)$ 
and $\psi(t)$ sampled in the same manner. The model $\pi$ takes in as input $M_{t_a}$ and 
$S_{t_a-t_\mathrm{model}}$ and produces as output $\tilde{S}_{t_a}$. $\tilde{S}_{t_a}$ 
differs from $S_{t_a}$ in that the former is the model's prediction whereas the latter is 
the ground truth. Taking everything into account, we obtain the following learning objective 
\begin{equation}
\begin{aligned}
        \argmin_{\Theta}\sum_{T_i \in D}\sum_{t_a \in (t_1, ..., t_k)}\sum_{i=0}^{\frac{t_\mathrm{pred}}{t_\mathrm{model}}-1}
        {||\pi(\tilde{S}_{t_a+t_\mathrm{model} \cdot (i-1)}, } \\
        {M_{t_a+i \cdot t_\mathrm{model}}) - S_{t_a+i \cdot t_\mathrm{model}}||_2^2} \label{learning_objective}
\end{aligned}
\end{equation}
where $\Theta$ is the parameter set of $\pi$ and $\tilde{S}_{t_a-t_\mathrm{model}} = S_{t_a-t_\mathrm{model}}$ for $i=0$. 

\section{Optimization System Architecture}

\begin{figure}
        \includegraphics[width=8.5cm, height=4.5cm]{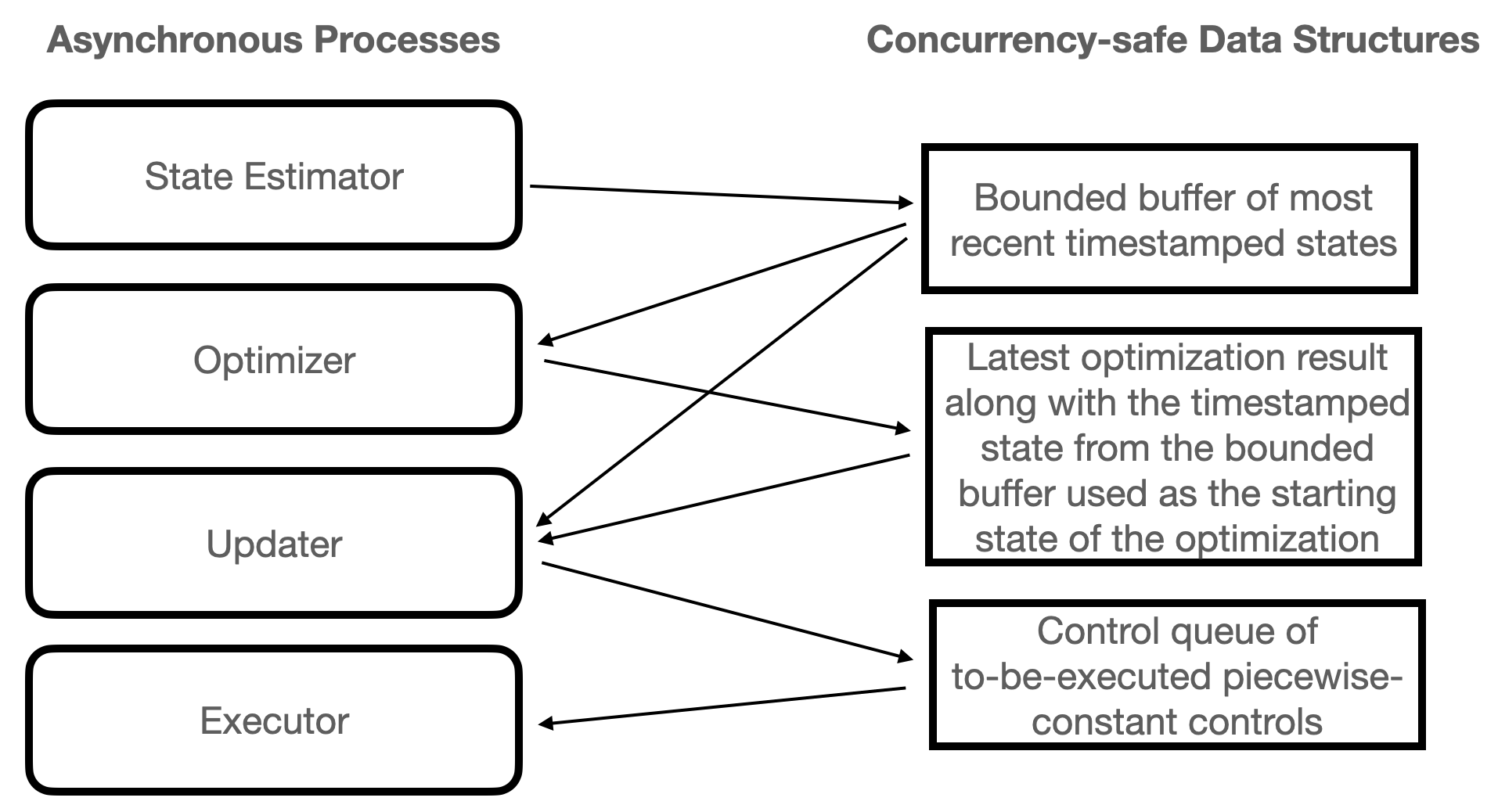}
        \caption{System architecture block diagram.}
        \label{block_diagram}
\end{figure}

In this section, we describe the system architecture of the proposed approach. We discuss how to integrate the 
optimization procedure and calls to the FKD model in a manner that enables real-time control 
on real robot hardware. Figure \ref{block_diagram} shows a block diagram of the system components. There are 
four key components that all operate asynchronously: the state estimator, optimizer, updater, and 
executor. 

\subsection{State Estimator}

In order for a predefined path to be followed accurately or for a goal state to be reached as fast as 
possible, accurate state estimates of the robot are essential. These state estimates define the robot's 
position, orientation, velocity, and angular velocity with respect to some coordinate frame. The 
state estimates themselves can come from a variety of sources such as a LIDAR based 
localization algorithm or visual odometry. It is assumed that there exists some delay $\epsilon$ between 
the actual, real-world state of the robot and output of the state estimator. The state estimator 
operates asynchronously, continuously updating a fixed sized buffer of the most recent state estimates. 
It is critical that each estimated state in the buffer is timestamped. 

\subsection{Optimizer}

The optimizer begins by obtaining from the state buffer with the most recent estimated state of 
the robot $\xi$. This state is used as the start state in the optimization procedure. Depending 
on the optimization objective, different steps are taken. For the path following objective, 
it is critical to first localize the robot on the map that the robot is desired to follow. Let $P$ 
be the path that the robot is assigned to follow at speed $v_\mathrm{desired}$. $P$ consists of 
a series of robot positions $x, y, \theta$ ordered in increasing order by which each position is to be 
reached by the robot. Each optimization will plan the next $\Delta t$ controls 
for the robot. To do this, the goal state $g \in P$ the robot needs to reach after time $\Delta t$ 
needs to be determined. Localizing the robot in $P$ amounts to finding the position $s \in P$ that 
minimizes $||\xi - s||_2^2$. The goal state $g$ can then be obtained by computing $g = P 
\rightsquigarrow v_\mathrm{desired} \cdot \Delta t$ where the $a \rightsquigarrow b$ operator looks ahead in $P$
from position $a$ by $b$ distance. 

Next, the optimizer prepares the input required by the forward kinodynamic model, namely the past 
$t_\mathrm{model}$ units time of robot state information. This is done by running time synchronization 
on the states in the state buffer by making use of the states' timestamps to sample $\frac{t_\mathrm{model}}
{\tau}$ evenly spaced states. Finally the optimization procedure is called, which optimizes over the next $\Delta t$ of controls by making $\frac{\Delta t}{t_\mathrm{model}}$ calls to the 
FKD model. The result $u^*$ along with $\xi$ is stored in a concurrency-safe data structure. Note that the optimization horizon $\Delta t$ differs from the prediction horizon $t_\mathrm{pred}$ in equation~\ref{learning_objective}.

\subsection{Updater}

The main role of the updater is dealing with latencies that are characterstic of real robot systems. 
We consider the two most impactful latencies: $\epsilon$, the previously introduced latency in the 
state estimators measurements, and $\gamma_i$, the time required for optimization $i$ to 
complete. Since $\gamma_i$ is different for every run of the optimizer, it needs to first be computed. 
This is done by computing the difference between the current system time and the timestamp of $\xi$ that 
was used as the starting state of the optimization procedure. The first $\gamma_i + \epsilon$ time units 
of controls are discarded from $u^*$, and the remainder replaces the contents of the control buffer. 

\subsection{Executor}

Finally, the executor asynchronously executes the commands stored in the control queue one at a time 
on the robot. 
Given a well-trained and accurate forward kinodynamic model $\pi$ the evolution of the state of 
the robot after executing the controls will closely match the predicted state sequence by the model, meaning 
the controls output by the optimization procedure are the desired ones.

\section{Experiments}

\begin{figure}
        \includegraphics[width=8.5cm]{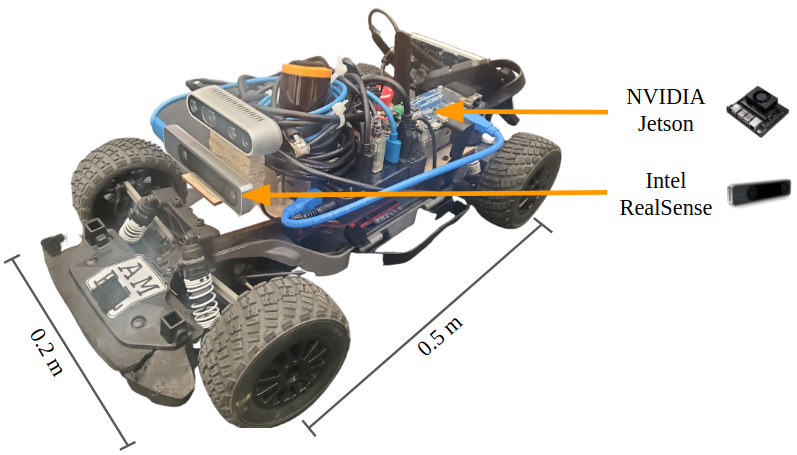}
        \caption{UT-Automata F1Tenth Robot Car 
        }
        \label{ut_automata}
\end{figure}

To evaluate the performance of our proposed Optim-FKD approach, we perform two sets of experiments, 
each involving a different variant of the robot control problem. We demonstrate the ability of 
Optim-FKD to successfully complete both, and show improved performance over an optimization-free, IKD 
model baseline~\cite{xiao2021learning}. 

\subsection{Experimental Setup}

\begin{figure*}
    \centering 
    \begin{subfigure}{0.25\textwidth}
        \centering 
        \includegraphics[width=4.25cm]{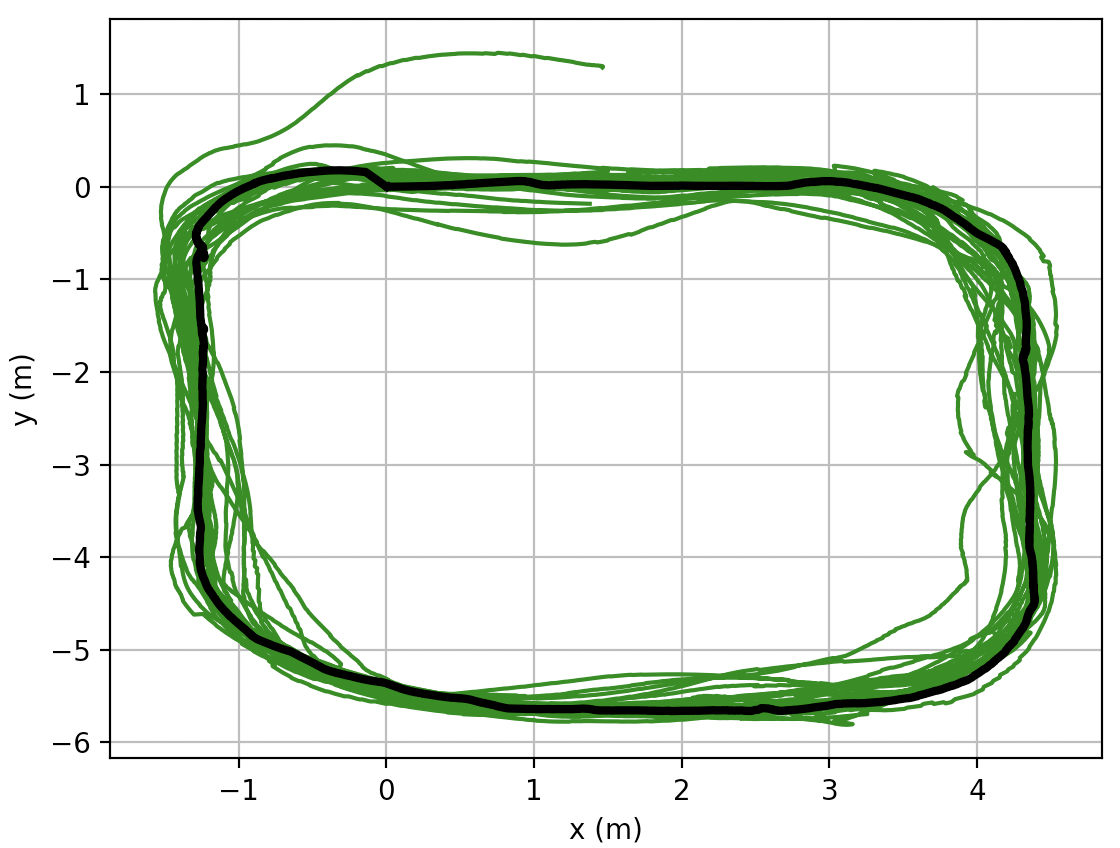}
        \caption{Optim-FKD on rounded rectangle}
    \end{subfigure}%
    \begin{subfigure}{0.25\textwidth}
        \centering 
        \includegraphics[width=4.25cm]{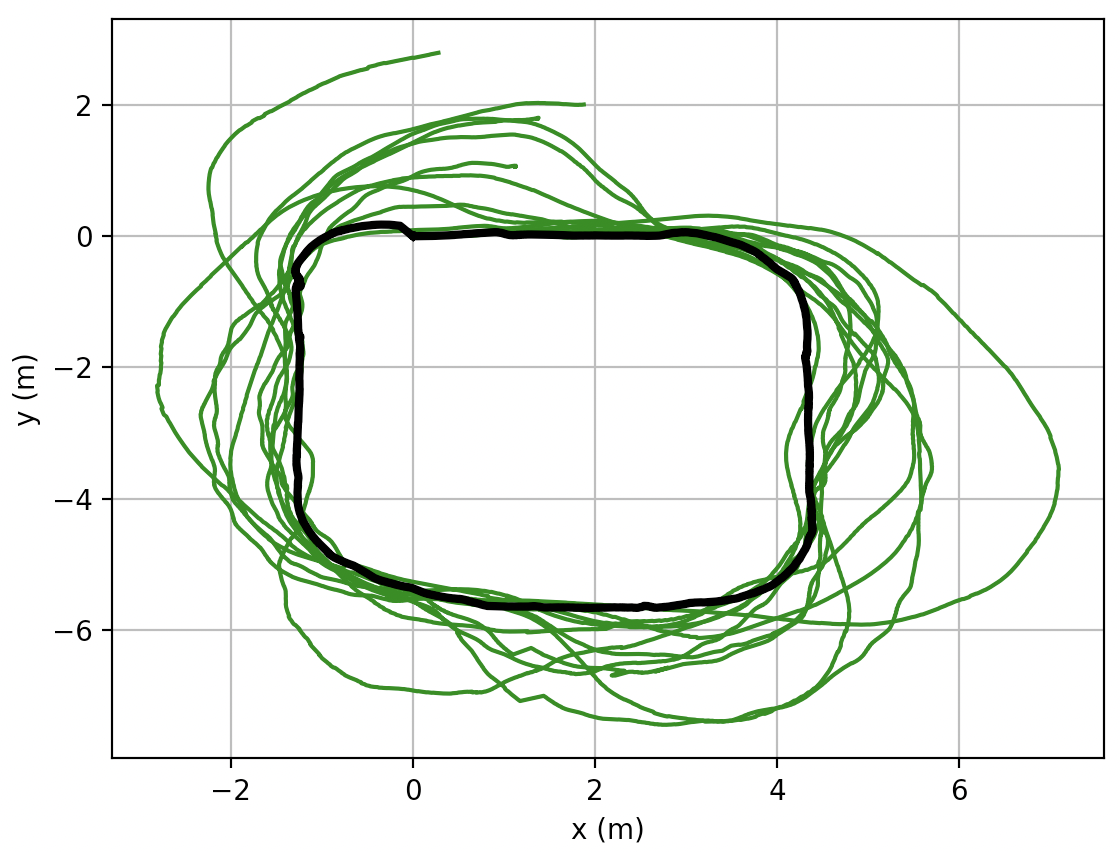}
        \caption{IKD on rounded rectangle}
    \end{subfigure}%
    \begin{subfigure}{0.25\textwidth}
        \centering 
        \includegraphics[width=4.25cm]{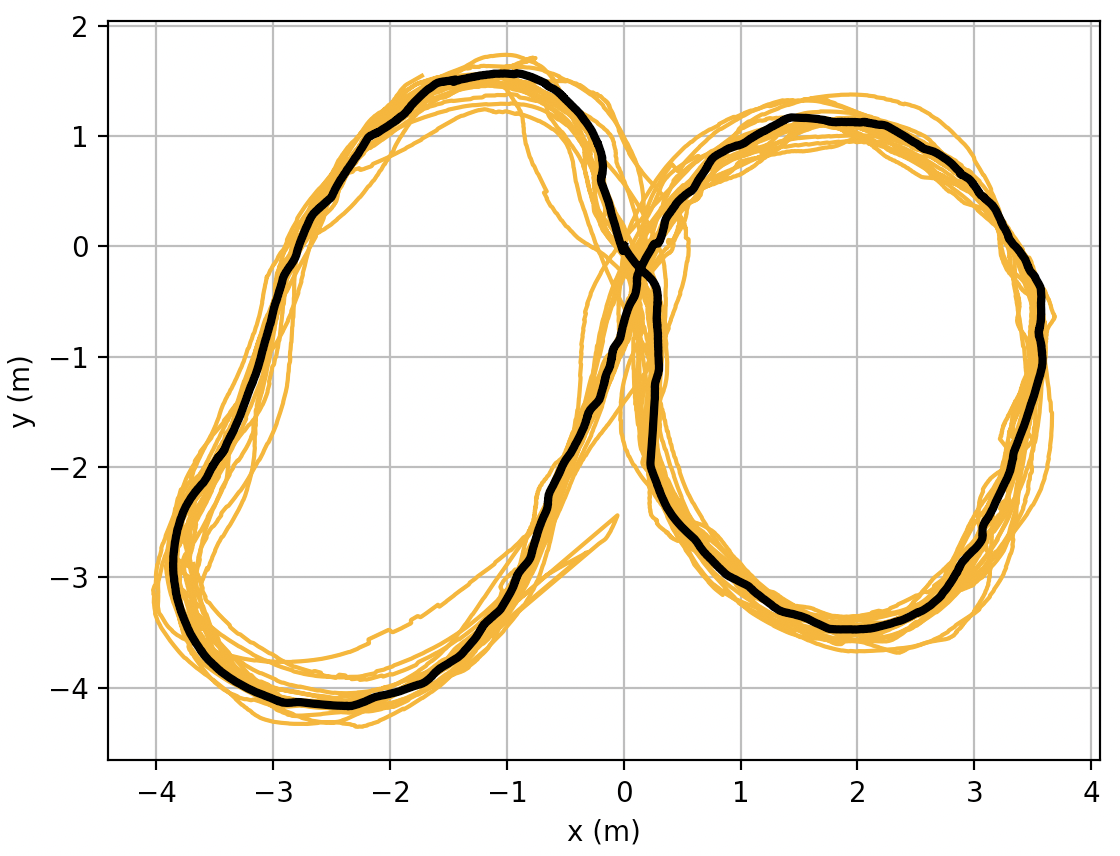}
        \caption{Optim-FKD on figure 8}
    \end{subfigure}%
    \begin{subfigure}{0.25\textwidth}
        \centering 
        \includegraphics[width=4.25cm]{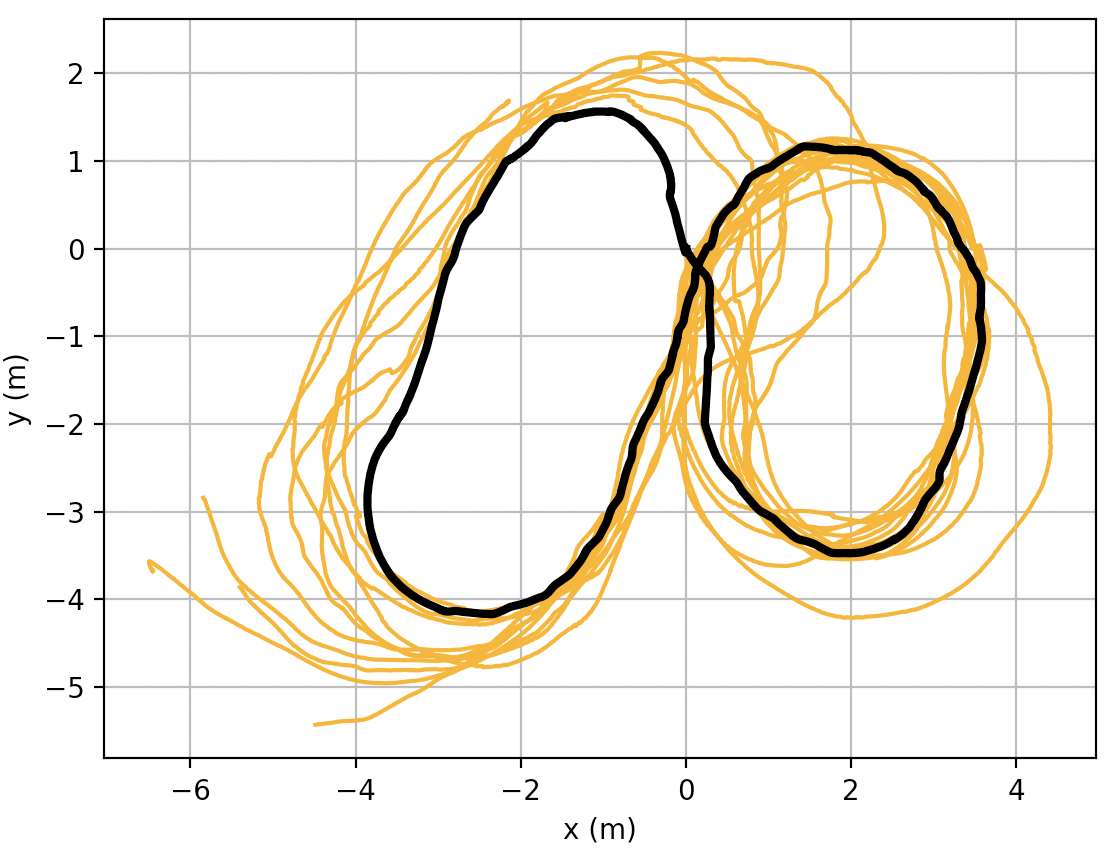}
        \caption{IKD on figure 8}
    \end{subfigure}%
    \caption{Superposition of execution traces of both the Optim-FKD and IKD algorithms running at different speeds on the rounded rectangle and figure 8 paths. In (a) and (c), which correspond to Optim-FKD, the execution trace matches the desired path (black) very closely. In contrast, there are significant errors in the baseline IKD model shown in (b) and (d).}
    \label{path_traces}
\end{figure*}

\begin{figure*}
    \centering 
    \begin{subfigure}{0.5\textwidth}
        \centering
        \includegraphics[width=8.5cm]{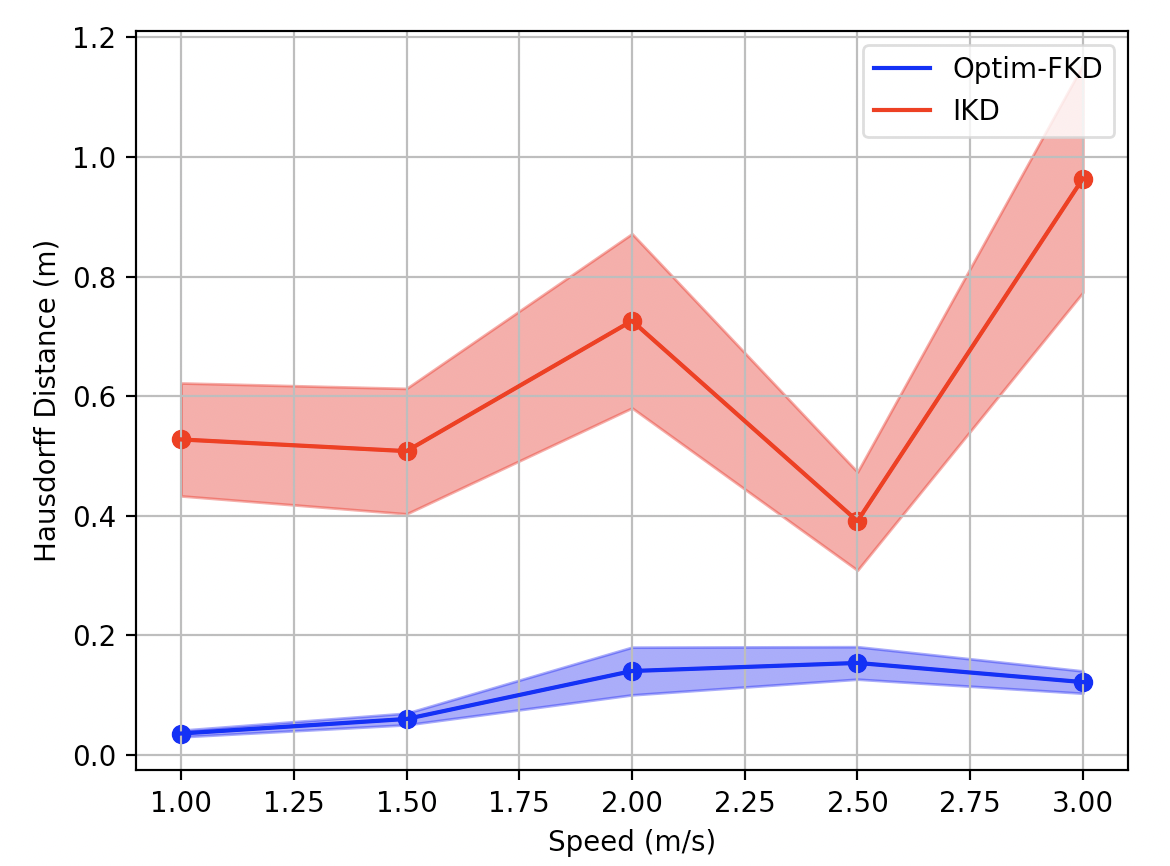}
        \caption{Rounded rectangle path}
    \end{subfigure}%
    \begin{subfigure}{0.5\textwidth}
        \centering
        \includegraphics[width=8.5cm]{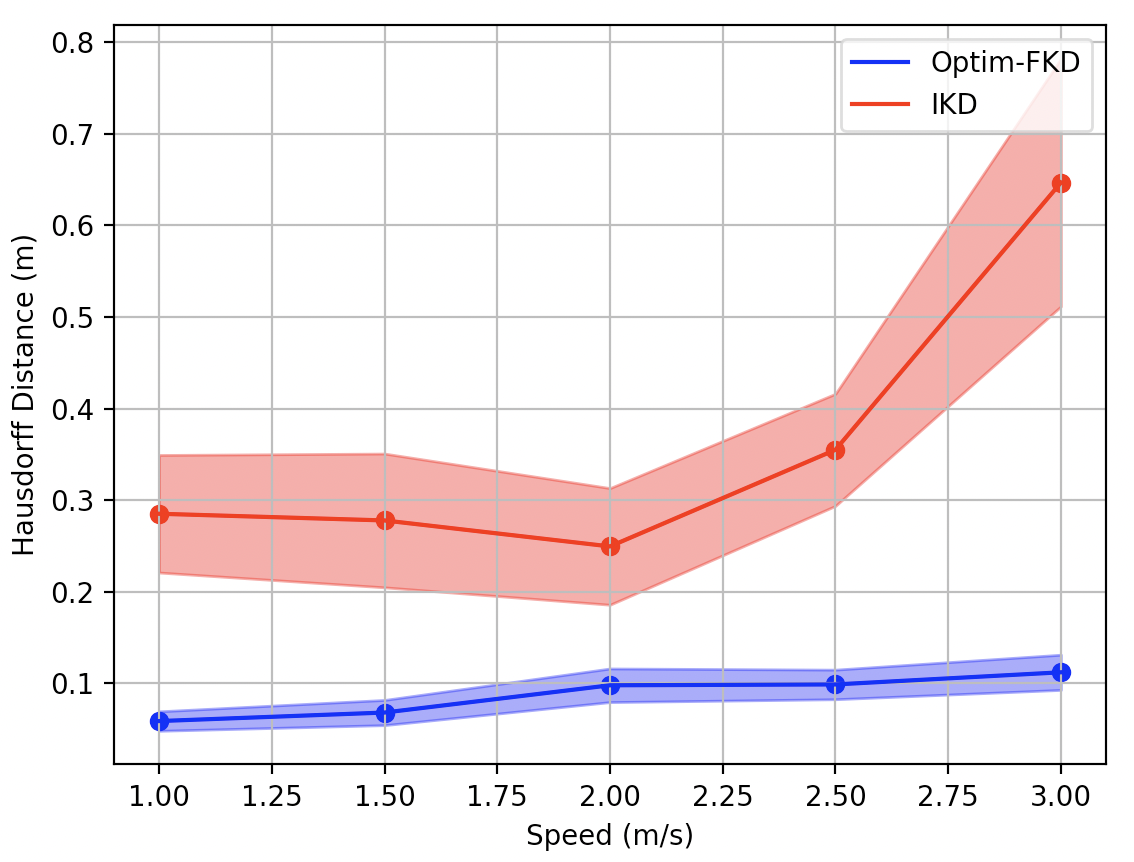}
        \caption{Figure 8 path} 
    \end{subfigure}%
    \caption{Average Hausdorff distances between the executed paths and the desired paths across different speeds. Each speed's average is computed for five rollouts.}
    \label{hausdorff_dists}
\end{figure*}

The robot platform used for our experimentation is the UT-Automata F1tenth Car depicted in figure 
\ref{ut_automata}. We make use of the car's Intel Realsense T265 tracking camera for localization and its Nvidia Jetson TX2 for compute. 

We consider two different experimental setups, each involving a different form of the robot control 
problem and thus a different optimization objective. For the first task we consider having the robot 
follow two different predefined paths as accurately as possible at speeds varying from $1.0~m/s$ to 
$3.0~m/s$. This task corresponds to the optimization objective presented in equation \ref{optim_objective_1}. For the second task we have the robot traverse from an initial starting state to a goal state in as little time as possible, without explicitly constraining which path it needs to take to reach the goal state. This task corresponds to the optimization objective presented in Eq. \ref{optim_objective_2}.

The baseline algorithm we compare against in all of the experiments is an inverse kinodynamic model. In contrary to the forward kinodynamic model, which maps a sequence of controls to the most likely next sequence of states, the inverse kinodynamic model attempts to infer what controls lead to a particular state sequence. To ensure fairness in the comparison, the underlying feedforward neural network used in both the FKD and IKD models was identical, with the same architecture ($6$ layers, $256$ neurons per layer) and activations (ReLU); the dataset that both were trained on was identical; and finally the number of iterations each was trained for was identical.

\subsection{Experiment 1: Path Following}

Here we assess the control capabilities of the Optim-FKD model on the task of accurately following a path. Figure \ref{path_traces} shows a visualization in black of the two paths used: a rounded rectangle path and a figure 8 path. Both paths were generated by joysticking the robot at a slow speed and recording only the position estimates. Each algorithm is assessed on how accurately it can follow the paths both at slower and higher speeds. For each speed, five full traversals through the desired path are completed. We measure an algorithm's ability to closely follow a path at a particular speed by computing the Hausdorff distance between the executed path and the desired path. 

Figure \ref{hausdorff_dists} shows the evaluation of the Hausdorff distance metric for (a) the rounded rectangle path and (b) the figure 8 path. Optim-FKD produces execution traces that are significantly closer to the desired trajectories than the IKD model. For both paths, the Optim-FKD algorithm is strictly under $0.2m$ Hausdorff distance, while the IKD algorithm is strictly greater than $0.2m$. Figure \ref{path_traces} provides a visual argument for these results: in (a) and (c) the executed path hugs the desired path much more closely than in (b) and (d). For both algorithms and across both paths, we observe that increases in velocity tend to result in higher Hausdorff distances. This is expected, since at high velocities the kinodynamic responses of the robot with respect to the terrain become more pronounced, making accurate control more difficult. 

We posit that the better performance of the Optim-FKD approach is explained by an inherent advantage in FKD models over IKD models: lesser impact of domain shift. Most robots where kinodynamic interactions with the environment are important are underactuated, i.e., the dimensionality of the control space is smaller than that of the state space. For a FKD model and an IKD model trained on the same dataset, the FKD model will learn a mapping from a lower dimensional space (the control space) to a higher dimensional space (the state space), whereas the IKD model learns the opposite mapping. During training, both the FKD model and the IKD model will be exposed to a limited subset of their input spaces - namely the space of teleoperable control inputs and the space of kinodynamically feasible state sequences. However when deployed, it is not guaranteed that the FKD model will receive control inputs characteristic of those during training, or the IKD model will be given kinodynamically feasible trajectories. Both models will suffer a domain shift, but the impact will be less for the FKD model. With the control space's lower dimensionality, the same quantity of training data will allow for the FKD model to have higher input space coverage than the IKD model, meaning there is less of a domain difference between train and test data. 

\subsection{Experiment 2: Optimal Connectivity}

\begin{figure}
    \centering
    \includegraphics[width=0.9\linewidth]{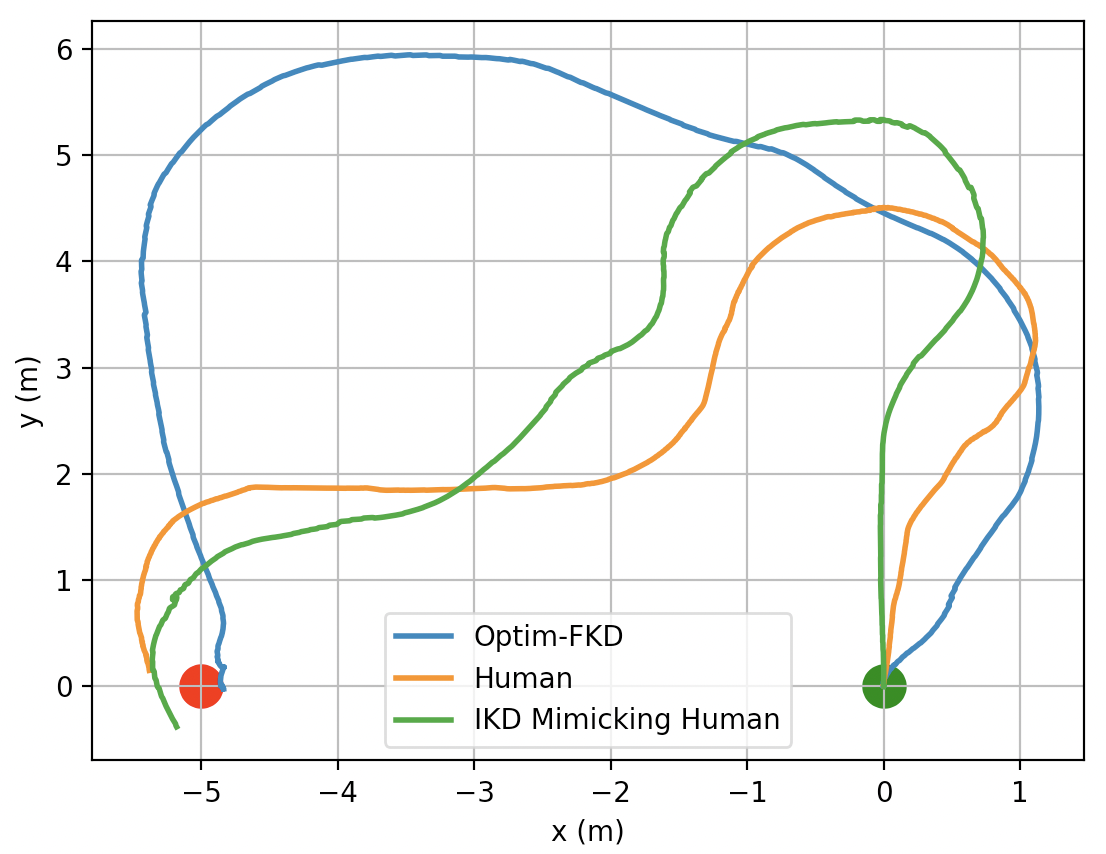}
    \caption{Paths produced by Optim-FKD, a human, and the IKD model following the human's path. Objective is to reach from the start (green) state at $(0, 0)$ to the goal (red) state at $(-5, 0)$ in minimum time. The start and end velocities are constrained to be $0~\frac{m}{s}$, the start heading is $\frac{\pi}{2}~rad$, and the end heading must be $\frac{3\pi}{2}~rad.$
    }
    \label{experiment_2_fig}
    
\end{figure}

In this experiment, we assess the generalizability of Optim-FKD compared to the IKD approach. As demonstrated in section \ref{formulation_two}, the optimal connectivity problem can be solved by Optim-FKD simply by altering the optimization objective. However, solving the same problem with the IKD approach requires the retraining of a new IKD model. This is because the IKD model has a fixed prediction horizon. Given a nearby state, it can output what controls to execute during this prediction horizon to reach the goal state. However in the general optimal connectivity problem, it may be possible that the goal state cannot be reached within the IKD model's prediction horizon, simply because it is too far away. Thus multiple evaluations of the IKD model would be necessary. But to do any one evaluation, the IKD model must know what the desired state is at time equal to its prediction horizon. This effectively means that the IKD model needs to be supplied with the so called racing lines, i.e., a user-specified path to take to reach the goal state. However, user-specified racing lines are prone to being suboptimal.

Figure \ref{experiment_2_fig} depicts the problem of reaching the goal position $(-5, 0)$ from the start position $(0, 0)$ in as little time as possible. Since, this task cannot be completed within the duration of the IKD model's prediction horizon, racing lines shown in yellow are provided by a human demonstrator. The IKD model attempts to trace the racing line as fast as possible shown. In contrast, Optim-FKD  runs the optimization described in Eq. \ref{optim_objective_2}, and is thus not dependent on racing lines nor requires a new FKD model to be trained. 

Using Optim-FKD, the robot is able to reach the goal state in $3.3~s$, whereas it takes $6.3~s$ for the robot to execute the path when using the IKD model. Thus Optim-FKD is able to find a more optimal path than the human provided one. Figure~\ref{experiment_2_fig} depicts the path taken by the robot under each navigation algorithm. Using Optim-FKD the robot makes a wider turn, which allows it to reach velocities higher than $3~\frac{m}{s}$ whereas the maximum linear velocity of the robot when using the IKD model is $2.6 \frac{m}{s}$

\section{Conclusion}

In this work we presented Optim-FKD, a new approach for accurate, high-speed robot 
control. We showed that solutions to various formulations of the robot control problem 
can be naturally expressed as a nonlinear least squares optimization with a FKD model. We 
demonstrate how such an FKD model can be learned effectively and integrated with the optimization 
on real robot hardware. Finally we evaluate our proposed approach on two robotic control tasks at high speeds and show that it outperforms the baseline. 

\section*{ACKNOWLEDGMENT}

This work has taken place in the Autonomous Mobile
Robotics Laboratory (AMRL) at UT Austin. AMRL research is supported in part by
NSF (CAREER-2046955, IIS-1954778, SHF-2006404), ARO
(W911NF-19-2-0333, W911NF-21-20217), 
DARPA (HR001120C0031), Amazon, JP Morgan, and Northrop Grumman Mission Systems.
The views and conclusions contained in this document are those of the authors
alone.

\bibliographystyle{IEEEtran}
\bibliography{references}

\addtolength{\textheight}{-12cm}   

\end{document}